\documentclass[a4paper, 10pt, conference]{ieeeconf}      

\IEEEoverridecommandlockouts                              
\usepackage{graphicx} 
\usepackage{listings}
\usepackage{color}

\definecolor{gray}{rgb}{.6,.6,.6}
\definecolor{purple}{rgb}{0.65, 0.12, 0.82}

\lstdefinelanguage{json}{
  comment=[l]{//},
  morecomment=[s]{/*}{*/},
  commentstyle=\color{gray}\ttfamily,
  stringstyle=\color{black}\ttfamily,
  morestring=[b]',
  morestring=[b]"
}

\title{\LARGE \bf
Work in Progress: Enabling robot device discovery through robot device descriptions
}

\author{Monica Anderson, Paul Kilgo, Chris Crawford and Megan Stanforth
\thanks{This work was supported by NSF grant CNS-1042360 and EEC-1005191.}
\thanks{M. Anderson, C Crawford and P Kilgo are with the Computer Science Department, The University of Alabama.
         Tuscaloosa, AL 35487 USA
        {\tt\small ccrawford@ua.edu, anderson@cs.ua.edu}}%
\thanks{M. Stanforth is with the Department of Computer Science, Florida State University,
        Tallahassee, FL 32306, USA
        {\tt\small megan@stanforth.com}}%
}

\begin{document}

\maketitle
\thispagestyle{empty}
\pagestyle{empty}

\begin{abstract}

There is no dearth of new robots that provide both generalized and customized platforms for learning and research.  Unfortunately as we attempt to adapt existing software components, we are faced with an explosion of device drivers that interface each hardware platform with existing frameworks.  We certainly gain the efficiencies of reusing algorithms and tools developed across platforms but only once the device driver is created.

We propose a domain specific language that describes the development and runtime interface of a robot and defines its link to existing frameworks.  The Robot Device Interface Specification (RDIS) takes advantage of the internal firmware present on many existing devices by defining the communication mechanism, syntax and semantics in such a way to enable the generation of automatic interface links and resource discovery.  We present the current domain model as it relates to differential drive robots as a mechanism to use the RDIS to link described robots to HTML5 via web sockets and ROS (Robot Operating System).

\end{abstract}

\section{INTRODUCTION}

Robot design deals with complexity in a manner similar to personal computers.  Robots have input/output devices that either provide output by acting in the environment or sensors that provide input.  Like PCs, robot peripherals contain firmware (device controllers) to predictably and efficiently manage resources in real-time.  Data is provided via a well-defined interface (set of system calls over a transport).  However, PCs abstract the differences in internal organization and chipsets through classifying devices in terms of their roles in the system.  These roles define an appropriate set of access and control functions that generally apply across the entire classification. Subsequent differences in devices are accommodated through the use of custom device drivers.

Robots also contain a mechanism for providing input and output to the higher-level algorithms, but the placement of the hardware abstraction layer is different than in personal computers.  Although most devices are classified according to the data type they produce and consume, classification occurs within the framework, not at the firmware level.  The disadvantage of this approach is that customized links from each hardware platform to each framework must be created.  In the current robotics landscape, this is a huge burden given the rate of innovation on new hardware platforms for many research and education purposes.  This ongoing backlog of creating one-to-one connections between platforms and hardware stifles innovation of control architectures.  The small number of developers comfortable with device driver creation either due to the unfamiliarity of the transports or the complexity of the threaded management of connections is source of slow progress.

Fortunately, we can leverage some commonalities found at the device driver level that link salient concepts both in the device driver domain and the robotics domain.  We propose a domain specific language based on these concepts called Robot Device Interface Specification (RDIS).  The RDIS describes the robot interface in terms of connection, primitives and interfaces.   An accurate characterization of the device domain enables some important innovations.  First, the RDIS enables a declarative, rather than a programmed interface to frameworks.   This approach benefits both device manufacturers and framework developers and users by separating the device semantics from the framework semantics.  Robot designers can describe the interface that they provide via onboard firmware and how it maps to abstract concepts via the RDIS.  The framework developers are only responsible for providing a mapping from the abstract concepts to the framework.   The abstract interface allows a many-to-many mapping between devices and frameworks using only a single map for each device and framework.  This is beneficial because manufacturers typically only provide device drivers for a single, often proprietary framework.  Specific device drivers for many frameworks are left to either framework developers (in the case of popular robots) or framework users (as needed).  The lack of available drivers for a specific device on a specific framework can be a barrier to  leveraging existing software components. 

Second, an abstraction that maps device semantics to domain specific concepts enables a new generation of development and runtime tools that can discover and manage available resources at both development and runtime.  Expertise in creating efficient threaded drivers for specific frameworks can be reused. This approach would simplify development by presenting developers with available resources that conform to specific domain concepts.

In this paper, we present the RDIS work in progress including RDIS specification and tools as well as a use of the RDIS to generate device specific programs.  The rest of this paper is organized as follows:  Section \ref{rw} discusses work related to declarative descriptions of robot hardware.  Section \ref{rdis} presents the preliminary domain model and its applicability to existing platforms.  The current implementation is discussed in Section \ref{case}.  The summary and future work directions are detailed in Section \ref{summary}.

\section{Related Works}
\label{rw}
Although the literature reveals very few attempts at using DSLs for hardware device drivers, Thibault et al \cite{Thibault} report the creation of efficient video device drivers using a novel DSL \cite{urbi}.   This language is targeted at the hardware interface layer and creates device driver code rather than interpreting code for efficiency.   URBI (Universal Robotic Body Interface) focuses on creating a model that controls the low level layer of robots and is independent from the robot and client system due to the client/server architecture.   Others \cite {lapham1999robotscript, bruyninckx2007towards} have attempted to address the lack of standardization in abstraction layers but have not considered moving abstractions to drivers using device descriptions.

Some frameworks use a declarative description of the robots for simulation.  Player/Stage ~\cite{gerkey01} is both a 2D simulator and a robot control framework.  Robot description files  are broken into two pieces: 1) a 2D description of the robot and its sensors and 2) a set of interfaces that abstract the data produced by hardware to a standard format.  The description, used for simulating the robot, consists of a polygon-based footprint with sensor locations marked. Actuation and sensor characteristics along with parameters for simplified error models are used to complete the model of the robot.  A domain-specific set of classes and message types describe what data can be obtained or how the robot can be manipulated (i.e. {\sc pose2d} for position  and {\sc laser} or {\sc ir} for distance to other objects).  The classes and message types represent the interface that abstracts the robot hardware to the data that it can produce or consume.  Writing software to the interfaces that a robot can utilize (rather than the specific robot) allows software to be written either for a simulated robot or a real robot, which in turns eases the transition from simulation to physical implementation.


ROS \cite{rospaper} targets a 3D simulation framework (Gazebo) and more sophisticated intelligent controller, which require a more rigorous description.  URDF (Uniform Robot Description Format) provides a 3D physical description broken into links and joints to facilitate not only mobile robots but manipulators as well.  Geometric bounding boxes and meshes allow for collision detection and realistic visualization.  Like Player/Stage, ROS utilizes a message-based model to decouple data providers from data producers.  Ideally robots that provide and consume similar data types can be controlled similarly.  Unlike Player Stage, URDF not only serves as a mechanism for simulating robots, but also allows for the visualization of real robots in both real-time and off-line (through saved messages). 

A select number of robot control frameworks move beyond visualization information and relevant interface declaration in the hardware description.  PREOP, an Alice-based programming interface \cite{tapia} for robots takes this paradigm further.  Not only is 3D visualization information supplied, but also the programming interface is completely specified by the selection of the robot object.  This is accomplished by linking the real-time control mechanism and exposed API available to the user within the robot object. 


\section{Robot Device Interface Specification (RDIS)}
\label{rdis}
\begin{figure*}[thpb]
      \centering
      \includegraphics[width=6in]{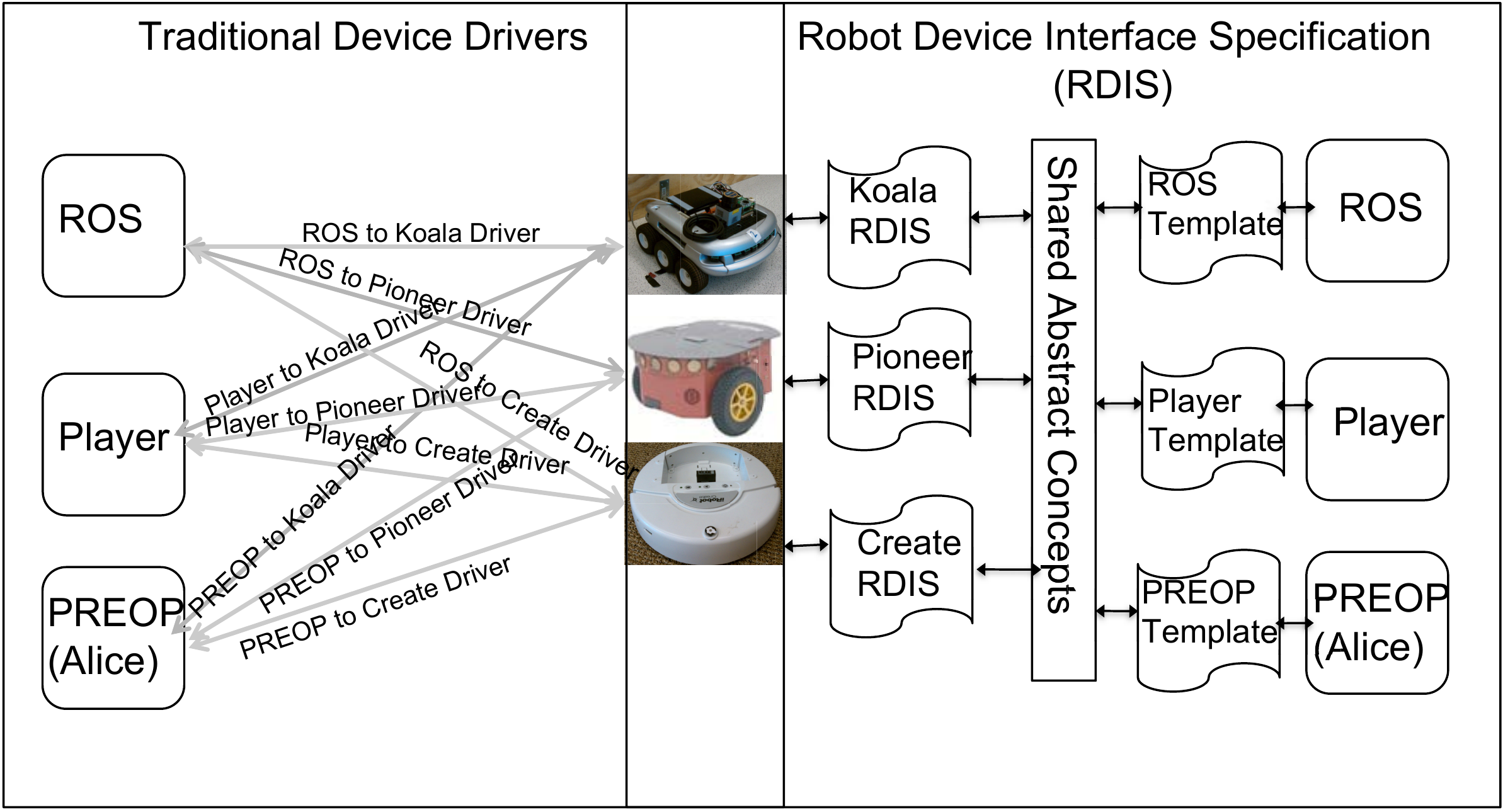}
      \caption{Traditional mappings between devices and frameworks utilize one to one mappings through device drivers.  We propose a separation of device semantics from framework semantics through the use of the RDIS to describe devices in an abstract manner.}
      \label{mapping}
\end{figure*}

\begin{figure*}[thpb]
      \centering
      \includegraphics[width=6in]{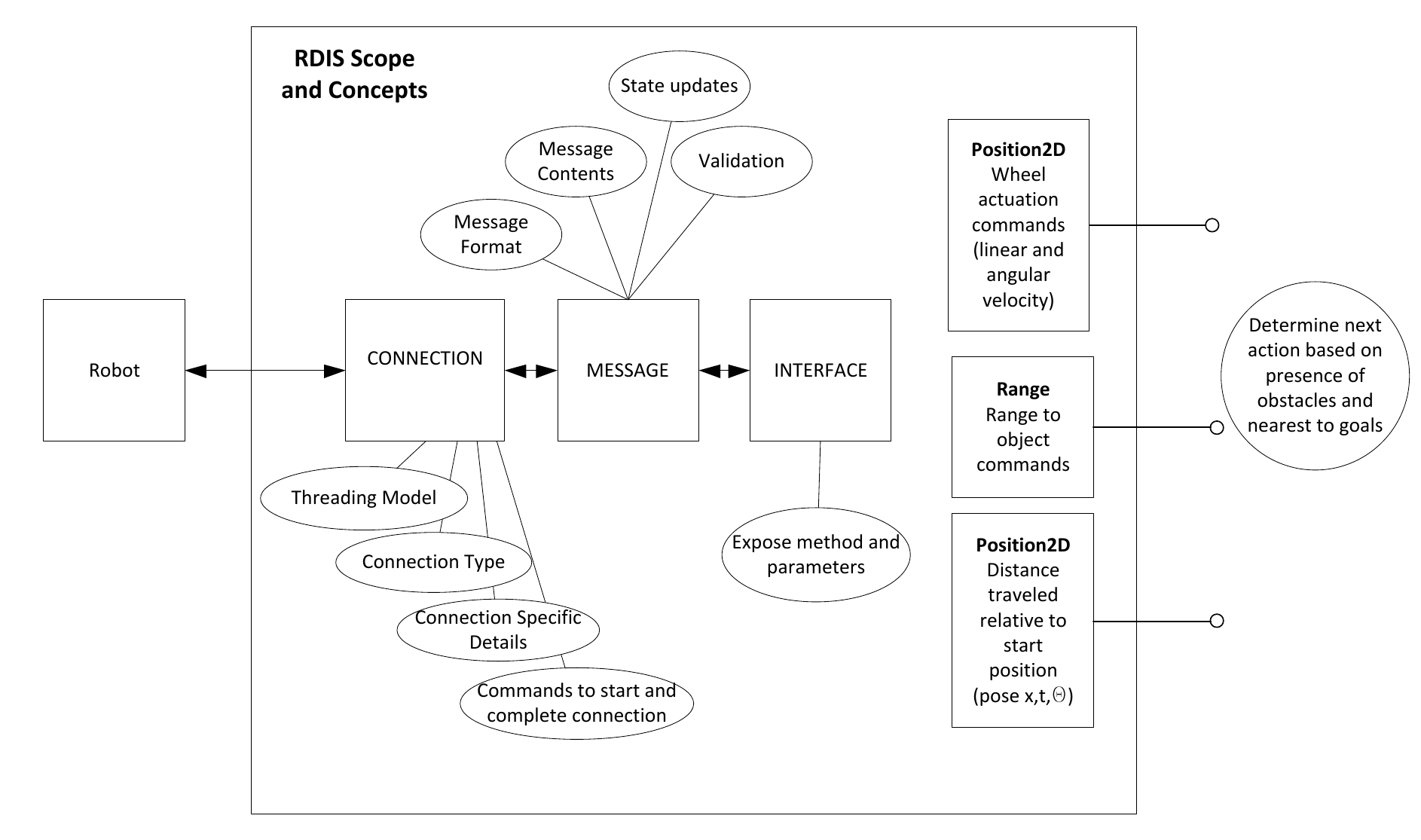}
      \caption{Diagram of RDIS concepts.}
      \label{er}
\end{figure*}

\begin{figure}[thpb]
      \centering
      \includegraphics[width=3in]{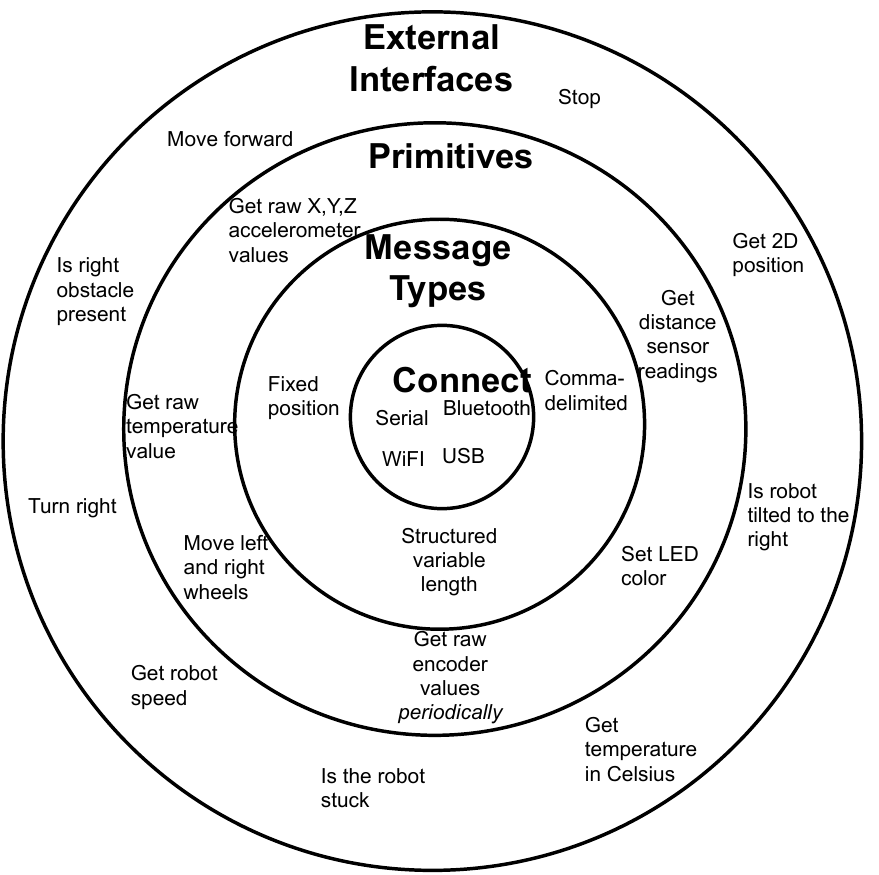}
      \caption{Relationship of concepts in typical device drivers or programs.}
      \label{circles}
\end{figure}
Frameworks and general reuse within the robotics research community rely upon the relatively invariant nature of mobile robots in several ways.  First, in an effort to reduce the complexity of control software, many robots reuse certain kinematic designs.  For example, differential drive is a fairly common choice as a configuration.  There is a computationally simple, closed form solution for forward and inverse kinematics and when combined with wheel encoders, provides a method for calculating pose relative to a starting position (which in turn enables closed loop control).  Although manipulators can contain arbitrary linkages, typically robots are constrained to configurations that provide a closed form inverse kinematic formulation and are numerically conducive to path planning (avoiding singularities) \cite{horn}.  Therefore, software that takes advantage of the kinematic control inherent in one configuration could be applied to other robots that reuse that configuration with appropriate parameterization.

Second, many robots including the popular Mobile Robots Pioneer class, iRobot Creates, K-Team robots, Erratic ER-1, White Box Robotics Model 914, Ar.Drones, and BirdBrain Finches contain an embedded firmware controller that accepts commands via a serial, Bluetooth, WiFI or USB interface rather than require the users to download a program to onboard memory.  This approach is popular because it allows the hardware designers to hide the complexity of hardware control within the firmware. There are a few designs that still expect developers to download code to the firmware.  The benefits of the low latency of local control are far outweighed by the burden of identifying a local toolchain to build the remote executables and the complexity of testing on a remote platform.  To that end, robots that utilize local control often provide modes where the local software program presents an API to an external computer (i.e. Lego Mindstorms via Lejos and E-Puck).

RDIS, Robot Device Interface Specification, is a domain specific language that defines the connection to robot firmware and maps data types to defacto standard messages for use in frameworks.  This mechanism provides an abstraction layer between the device and frameworks that negates the need for device drivers as point solutions.  The RDIS has three purposes: 1) provide enough information for simulation and visualization of hardware and controllers, 2) declaratively specify the mechanism for requesting data and actuation, and 3) inform users of standard message types that can be obtained from the hardware to facilitate connections to existing frameworks.  The RDIS enables several efficiencies in robotics controller development.  Although the long term goal is to embed the RDIS within the firmware as a response to a request, it could also be requested via the Internet from a repository.   However manufacturers that provide access to the RDIS within their hardware would benefit from being able to take advantage of the RDIS connectors available for frameworks without specifically providing device drivers.  Then the RDIS serves as a discovery message to the development architecture regarding the services available and how to manage the services at runtime.  Making the hardware the system of record for its abilities is in line with other modern technologies (Bluetooth for example).  The challenge in successfully defining the RDIS is in creating a model that captures the generalizable aspects of robots and appropriately identifies the aspects that vary.

Domain models, when designed properly, can be somewhat invariant to changes and can provide a stable basis for deciding the structure and parameters of the specification.  Primary concepts include connection, message formats, primitives and exposed interfaces (Figure \ref{er}).  Figure \ref{circles} shows a diagram of the domain and the scope of the RDIS.  Connections are generally through standard transports and describe how the robot connects to external controllers.  Message formats either encode parameters in ASCII formats or send natively as byte values. Primitives describe the device invariant features which are requests that can be made and ingoing and outgoing parameters.  Exposed interfaces describe a more convenient exposed interface that may map directly to primitives or may add value to a primitive by data conversion or specific parameterization.   There are some cross-cutting concerns.  Messaging paradigms are either request-reply (service-based and adhoc) or publish/subscribe (periodic updates that are published or expected).  Threading models include single (one loop that services incoming and outgoing data), dual threaded (one thread for servicing incoming requests and one thread for periodic requests), or multiple threads (requests create threads and periodic requests are on different frequencies).  Some drivers maintain state (i.e. current position relative to the starting point) and the validation routines for incoming data and read and write routines can vary.

\section{Case Study}

\label{case}
\begin{figure}[thpb]
      \centering
      \includegraphics[width=2.5in]{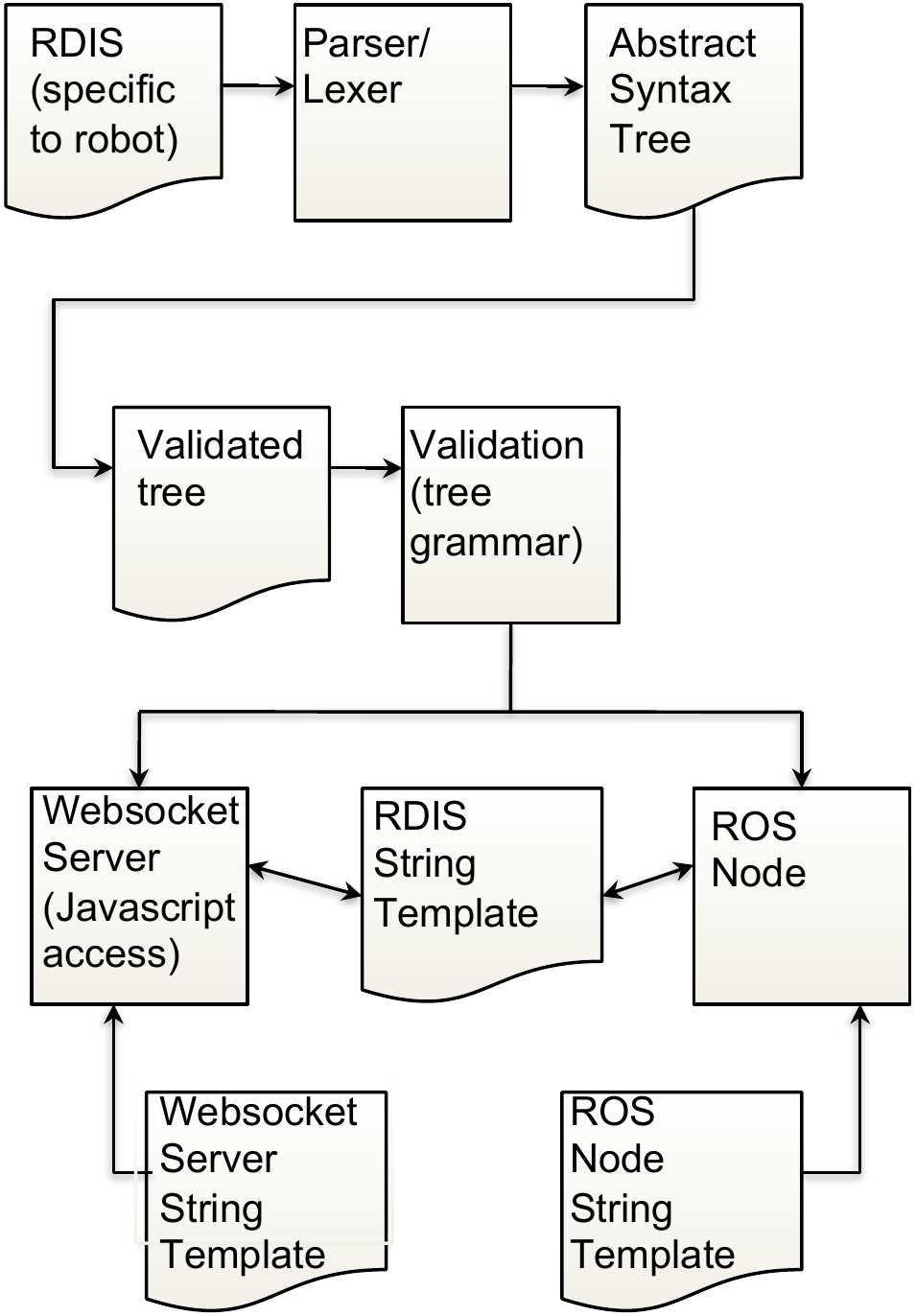}
      \caption{Steps to generate device specific programs and drivers from the RDIS.}
      \label{generate}
\end{figure}

\begin{figure*}[thpb]
      \centering
      \includegraphics[width=6in]{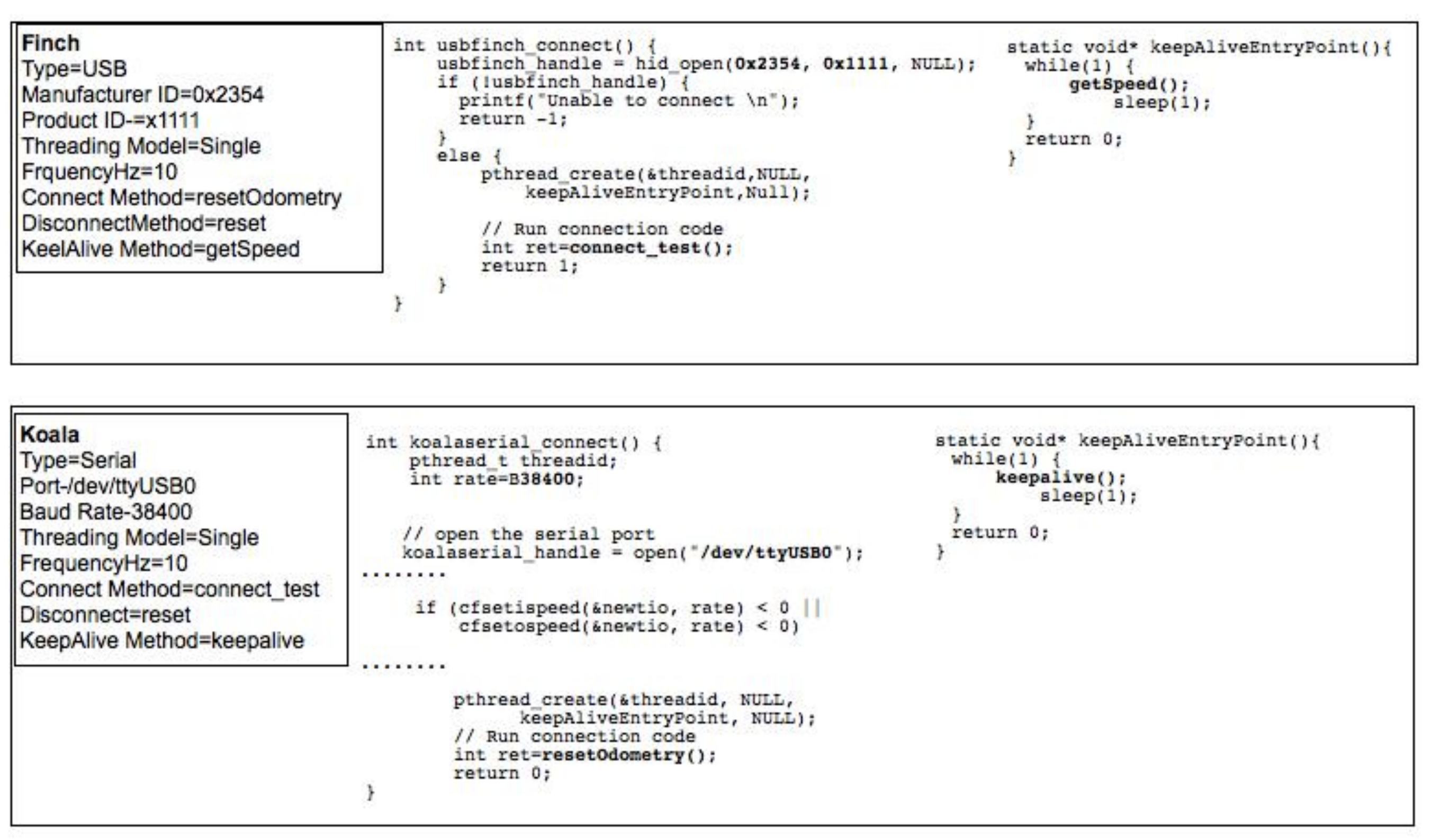}
      \caption{RDIS statements that support USB and serial connections.  Outside of connection parameters, the remainder of the statement describes the threading model and connection specific processes.}
      \label{connect}
\end{figure*}


A preliminary RDIS that meets requirements 2 and 3 has been implemented for the Finch robot from Bird Brain and the Koala from K-Team.  Figure \ref{generate} shows how the RDIS is used to create robot specific driver code for frameworks.  The RDIS and the resulting templates contain attributes to describe several functions including connection, basic primitives, external interfaces and mapping to abstract robotic concepts.  The connection statement delineates the physical connection parameters, the overall threading model and functions to call upon the creation of the connection (excerpt shown in Figure ~\ref{connect}).  Depending upon the physical connection, other parameters could include port ID, serial connection parameters, or USB ID.  Although we intend to support three threading models, the {\sc single} threading model is used which processes requests and publishing of data in a single active loop.  A callback is used to process any subscriptions (if supported by the framework and indicated by the abstract mapping) and second thread is used to issue a keepalive command if required by the platform.  All data protection, including appropriate mutexes are generated by the RDIS handler based on the threading model selected.
%
%
%
%
%
%

Basic primitives describe the mechanism for sending information to and from the robot.  Primitive specifications indicate the associated connection (described in connection statement), frequency and message formatting.  Frequency indicates whether the method is a request/reply or periodic.  Request/reply methods (indicated as adhoc) are only submitted when a request is received.  Parameters can be provided by the client and data can be returned to the calling client.  Periodic requests are executed on a schedule and utilize a set of state variables (defined in state variables section) to retrieve and save method data.  Message formats for communicating with robot firmware are either position based or delimited.  The interim specification presented here encodes the messages along with the input and output fields.  An example of the setMotor function in Figure~\ref{setmotorc} and the underlying abstract syntax tree is shown in Figure~\ref{setmotor}.

\begin{figure}[thpb]
	\centering
\begin{lstlisting}[language=c,breaklines=true,numbers=none]
int setMotors(int leftDir,int leftPower,int rightDir,int rightPower) {

	char buffer[9];
	memset(buffer,0,9);
	//format bytelist
	buffer[1]=(char)0;
	buffer[2]='M';
	buffer[3]=(char)leftDir;
	buffer[4]=(char)leftPower;
	buffer[5]=(char)rightDir;
	buffer[6]=(char)rightPower;

	int ret = usbfinch_write(buffer,9);
	return 0; 
}
\end{lstlisting}
	\caption{setMotor function generated from the setMotor RDIS definition.}
	\label{setmotorc}
\end{figure}

\begin{figure*}[thpb]
      \centering
      \includegraphics[width=6.75in]{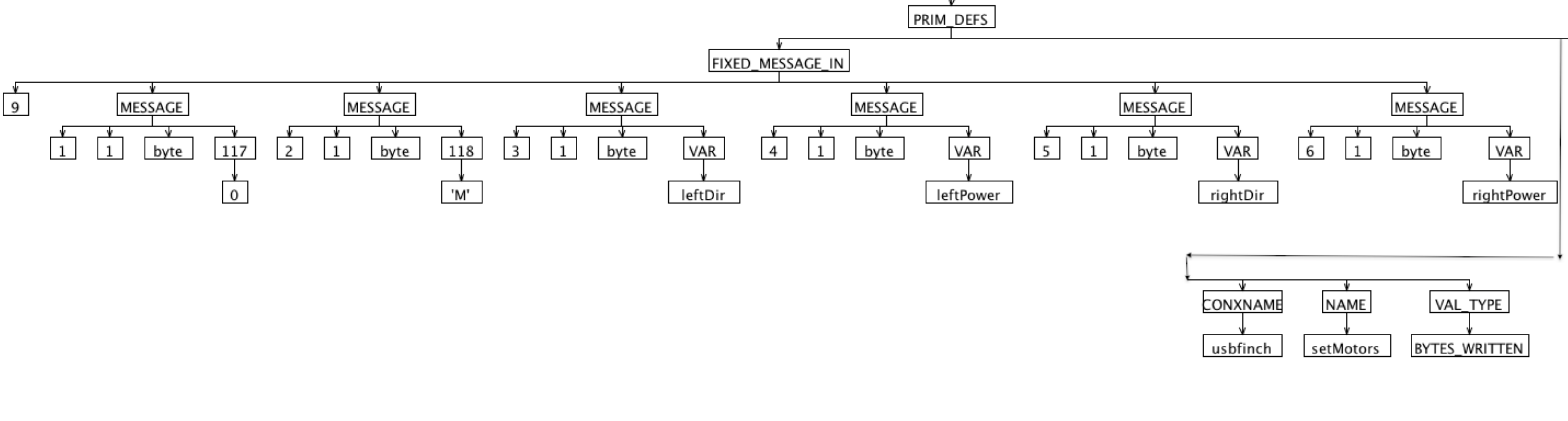}
      \caption{Abstract syntax tree for the setMotor function for the Finch.  It is used to generate the setMotor function shown in Figure~\ref{setmotorc}.}
      \label{setmotor}
\end{figure*}

The external interface exposes the API available to client programs.  Each interface is composed of one or more primitive methods or can return state variables (updated asynchronously by periodic primitives).   The separation between the external interface and primitives encapsulates the robot firmware and its parameters from developers.  For example, actuation commands are sometimes provided in encoder units where an external API would utilize a standard measure such as meters per second.   

\begin{figure}[thpb]
      \centering
      \includegraphics[width=3.5in]{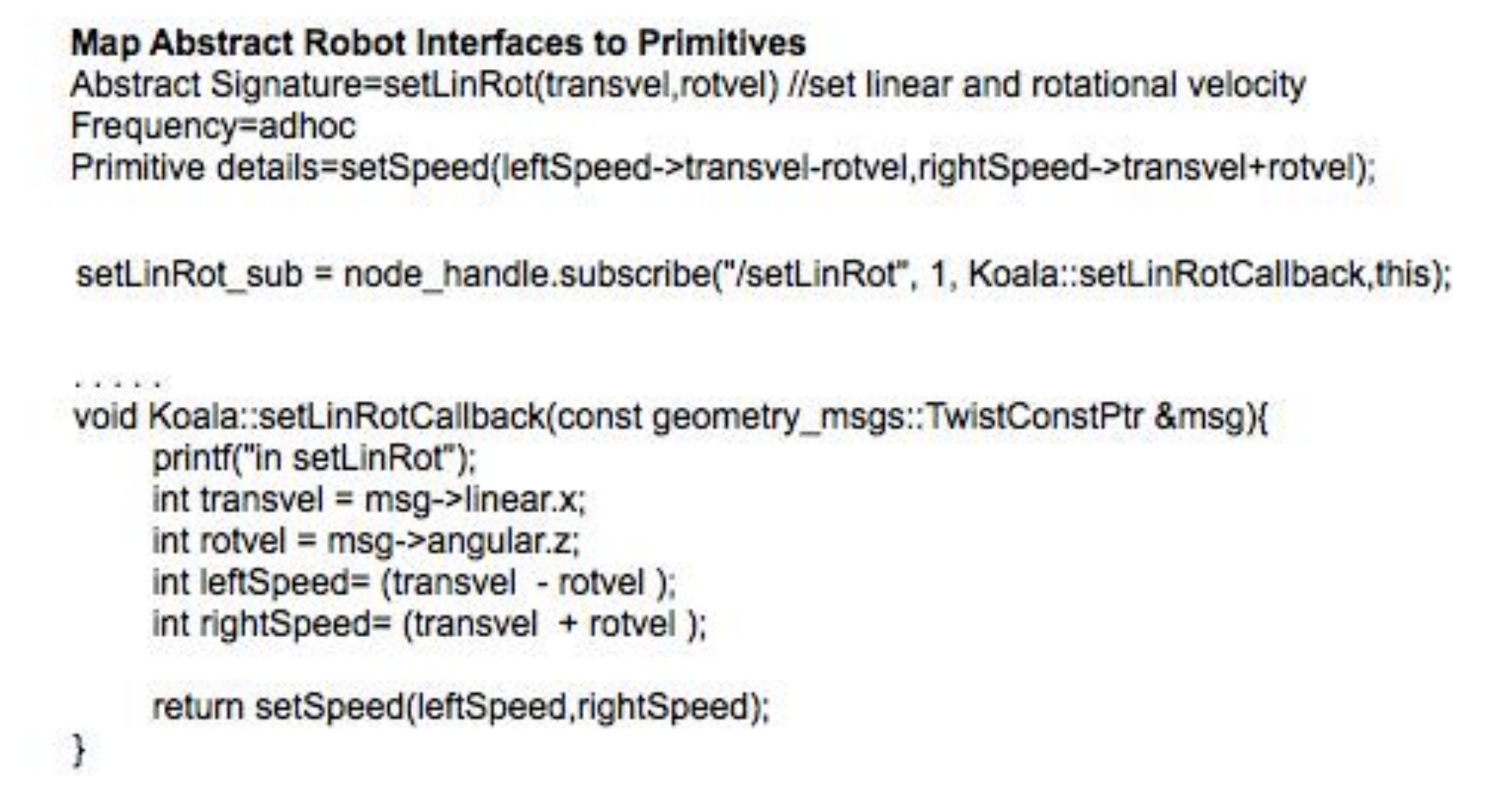}
      \caption{Abstract mapping links the external interfaces or primitives to abstract concepts utilized in many frameworks.  Building framework specific drivers requires that the templates map these abstract concepts to appropriate mechanisms within the framework.  Here the position2d interface which describes the movement of differential drive robots is mapped to the ROS twist message.}
      \label{am}
\end{figure}

A mapping to abstract concepts in sensing and locomotion provides a link between robots and existing frameworks.  Rather than specify framework specific information within the RDIS, abstract concepts that describe the data available are used instead.  For example, since a differential drive robot can be controlled via linear and rotational velocity, we provide a mapping between linear and rotational velocity and robot primitives (left and right velocity).  An example is shown in Figure ~\ref{am}.

In the current design, the RDIS is modeled as a JSON subset (excerpt in Figure~\ref{rdisJSON}).  The intermediate product is an abstract syntax tree that represents the robot details in a domain specific model.  This intermediate format can be further processed to verify conformance to the specification.  End products are generated from the verified syntax tree, either in a single or multiple passes, using templates that format data based on the model.  The preliminary result of this approach includes RDIS specifications and grammars that generate a command line program, websocket server and a ROS driver.  The ROS driver looks for specific interface signatures in the abstract mapping section that match to ROS message structures.

It is important to note that the RDIS toolset is enabled by ANTLR and StringTemplate.  These are open source libraries that parse and process data according to grammars.  These grammars are often used to define domain specific languages that are subsequently processed either by interpreters or translators.  The sample RDIS and the translation to a C-based command line controller and a websocket server and a C++ ROS node were achieved through the use of grammars and the ANTLR and StringTemplate libraries.  Although these libraries provide many built-in features, the ability to embed code to customize processing is important to using these tools effectively.

%
%


\begin{figure}[thpb]
	\centering
\begin{lstlisting}[language=json,breaklines=true,numbers=none]
{
  "name": "Koala",

  "connections": [{
    "name": "koalaserial",
    "type": "serial",
    "port": "USB0",
    "speed": "38400",
    "threadmodel": {
      "type": "single",
      "freqHz": "10"
    },
    "startprocess": "resetOdometry",
    "terminationprocess": "resetOdometry",
    "keepaliveprocess": "getCurrentSpeed"
  }],

  "primitives": [{
      "connection": "koalaserial",
      "name": "setSpeed",
      "format": "csv",
      "type": "in_primitive",
      "freq": "adhoc",
      "infieldconsts": [
        {"type": "string", "value": "\"D\""}
      ],
      "infieldvars": [
        {"type": "int", "value": "leftSpeed"},
        {"type": "int", "value": "rightSpeed"}
      ],
      "outfieldvars": [
        {"type": "string", "value": "cmd"}
      ]
  }],
  
  "interfaces": [{
      "name": "moveForward",
      "freq": "adhoc",
      "primitive": {
        "name": "setSpeed",
        "inparms": [
          {"name": "leftSpeed", "value": "220"},
          {"name": "rightSpeed", "value": "220"}
        ]
      }
    }],

  "position2d": {
    "setLinRot": {
      "name": "setSpeed",
      "inparms": [
        {"name": "leftSpeed", "value": "transvel-rotvel"},
        {"name": "rightSpeed", "value": "transvel+rotvel"}
      ]
    }
  }
}
\end{lstlisting}
	\caption{Excerpt of RDIS JSON syntax for the Koala robot.  The Koala uses a comma delimited string as a message format with a preliminary character to denote the command.  This format is different than the position formatted messages shown for the Finch in Figures~\ref{setmotor} and \ref{setmotorc}.}
	\label{rdisJSON}
\end{figure}

\section{Conclusion and Future Work}
\label{summary}
This preliminary result supports the idea that general robot devices can be described declaratively in a manner that supports discovery and that links to the backend processes.  The ultimate goal to enable more accessible programming by embedding the robot device descriptions within the device.  Discovery occurs when the design environment queries the device for its supported services (or APIs).  The initial approach for platforms that support onboard reconfigurable firmware is to augment the firmware to support a single command that communicates the RDIS.  The information provided by the RDIS can be used by any RDIS-enabled development environment.  It is expected that manufacturers will choose to RDIS enable their devices once there are more RDIS-enabled environments are available.

The RDIS must be expanded to be useful in a larger context.  These tasks include but are not limited to: 1) addition of a complete kinematic, visual and collision description consistent with existing simulators and frameworks, 2) error handling at both the communication and primitive levels, 3) implementation of additional threading models, 4) refinement of the state concept and how it matches to primitives and interfaces, 5) management of sensor and actuator error models consistent with existing frameworks, and 6) match internal mechanisms to framework standard interfaces and message types through linking the description and the exposed API instead of relying upon matching external interface signatures.  These changes require updates to the specification and the underlying parsers, lexers, tree grammars and string templates.

\section{ACKNOWLEDGMENTS}

The authors gratefully acknowledge the partial support NSF via grants CNS-1042360 and EEC-1005191.



\begin{thebibliography}{99}
\bibitem{urbi}
J.C. Baillie, Urbi: Towards a universal robotic low-level programming language, {\it Intelligent Robots and Systems, 2005.(IROS 2005). 2005 IEEE/RSJ International Conference on}, 820--825, 2005

\bibitem{lapham1999robotscript}
Lapham, J., RobotScript™: the introduction of a universal robot programming language,{\it Industrial Robot: An International Journal}, 26(2), 17--25, 1999

\bibitem{bruyninckx2007towards}
Bruyninckx, H.,Towards a Common Robotics, Automation and Manufacturing Operating System (CRAMOS), {\it System}, 2007

\bibitem{gerkey01}
B P Gerkey, R T Vaughan, K Stoy, A Howard, G S Sukhatme, M J Mataric ,Most Valuable Player: A Robot Device Server for Distributed Control
{\it Proc. of the IEEE/RSJ Intl. Conf. on Intelligent Robots and Systems (IROS)}, 2001

\bibitem{rospaper}
M. Quigley, B. Gerkey, K. Conley, J. Faust, T. Foote, J. Leibs, E. Berger, R. Wheeler, and A. Ng, "ROS: an open-source Robot Operating System," {\it International Conference on Robotics and Automation}, 2009.

\bibitem{alice}
S. Cooper, W. Dann, and R. Pausch, "Alice: a 3-D tool for introductory programming concepts," {\it Proceedings of the fifth annual CCSC northeastern conference on The journal of computing in small colleges}, pp. 107-116. 2000

\bibitem{tapia}
B.L. Wellman, J. Davis, and M. Anderson, "Alice and robotics in introductory CS courses," {\it The Fifth Richard Tapia Celebration of Diversity in Computing}, 2007
 
\bibitem{horn}
B. Horn, {\it Kinematics, Statics and Dynamics of Two-dimensional manipulators}
 
\bibitem{Thibault}
S.A. Thibault,   R.Marlet,   C.Consel,  Domain-specific languages: from design to implementation application to video device drivers generation, {\it IEEE Transactions on Software Engineering} Volume: 25 Issue: 3 ,  May/Jun 1999

\end{thebibliography}
\end{document}